\documentclass[conference]{IEEEtran}
\IEEEoverridecommandlockouts
\usepackage{cite}
\usepackage{amsmath,amssymb,amsfonts}
\usepackage{algorithmic}
\usepackage{graphicx}
\usepackage{textcomp}
\usepackage{xcolor}
\usepackage[colorlinks=true,linkcolor=blue,citecolor=blue,urlcolor=blue]{hyperref}
\usepackage{caption}
\def\BibTeX{{\rm B\kern-.05em{\sc i\kern-.025em b}\kern-.08em
    T\kern-.1667em\lower.7ex\hbox{E}\kern-.125emX}}

\makeatletter
\newcommand{\linebreakand}{%
  \end{@IEEEauthorhalign}
  \hfill\mbox{}\par
  \mbox{}\hfill\begin{@IEEEauthorhalign}
}
\makeatother

\begin{document}

\title{Multi-scale Frequency-Aware Adversarial Network for Parkinson's Disease Assessment Using Wearable Sensors\\
}

\author{\IEEEauthorblockN{Weiming Zhao}
\IEEEauthorblockA{\textit{National Pilot School of Software} \\
\textit{Yunnan University}\\
Kunming, China \\
zhaoweiming@stu.ynu.edu.cn}
\and
\IEEEauthorblockN{Xulong Wang}
\IEEEauthorblockA{\textit{School of Computer Science} \\
\textit{University of Sheffield}\\
Sheffield, UK \\
xl.wang@sheffield.ac.uk}
\and
\IEEEauthorblockN{Jun Qi}
\IEEEauthorblockA{\textit{Department of Computing} \\
\textit{Xi'an JiaoTong-Liverpool University}\\
Suzhou, China \\
jun.qi@xjtlu.edu.cn}
\linebreakand 
\IEEEauthorblockN{Yun Yang}
\IEEEauthorblockA{\textit{National Pilot School of Software} \\
\textit{Yunnan University}\\
Kunming, China \\
yangyun@ynu.edu.cn}
\and
\IEEEauthorblockN{Po Yang}
\IEEEauthorblockA{\textit{School of Computer Science} \\
\textit{University of Sheffield}\\
Sheffield, UK \\
po.yang@sheffield.ac.uk}
}

\maketitle

\begin{abstract}
Severity assessment of Parkinson's disease (PD) using wearable sensors offers an effective, objective basis for clinical management. However, general-purpose time series models often lack pathological specificity in feature extraction, making it difficult to capture subtle signals highly correlated with PD. Furthermore, the temporal sparsity of PD symptoms causes key diagnostic features to be easily ``diluted" by traditional aggregation methods, further complicating assessment. To address these issues, we propose the Multi-scale Frequency-Aware Adversarial Multi-Instance Network (MFAM). This model enhances feature specificity through a frequency decomposition module guided by medical prior knowledge. Furthermore, by introducing an attention-based multi-instance learning (MIL) framework, the model can adaptively focus on the most diagnostically valuable sparse segments.We comprehensively validated MFAM on both the public PADS dataset for PD versus differential diagnosis (DD) binary classification and a private dataset for four-class severity assessment. Experimental results demonstrate that MFAM outperforms general-purpose time series models in handling complex clinical time series with specificity, providing a promising solution for automated assessment of PD severity.
\end{abstract}


\section{Introduction}
Parkinson's disease (PD) is a common neurodegenerative disorder characterized by motor symptoms (such as resting tremor and bradykinesia) and non-motor symptoms that severely impact patients' quality of life\cite{b1}. Early and accurate diagnosis, along with continuous disease assessment, are crucial for optimizing treatment plans and slowing disease progression. Traditional clinical assessment relies mainly on the Unified Parkinson's Disease Rating Scale (UPDRS), which is time-consuming, susceptible to observer subjectivity, and struggles to capture daily symptom fluctuations\cite{b2}. In recent
\begin{figure}[htbp]
\centering
\includegraphics[width=0.48\textwidth]{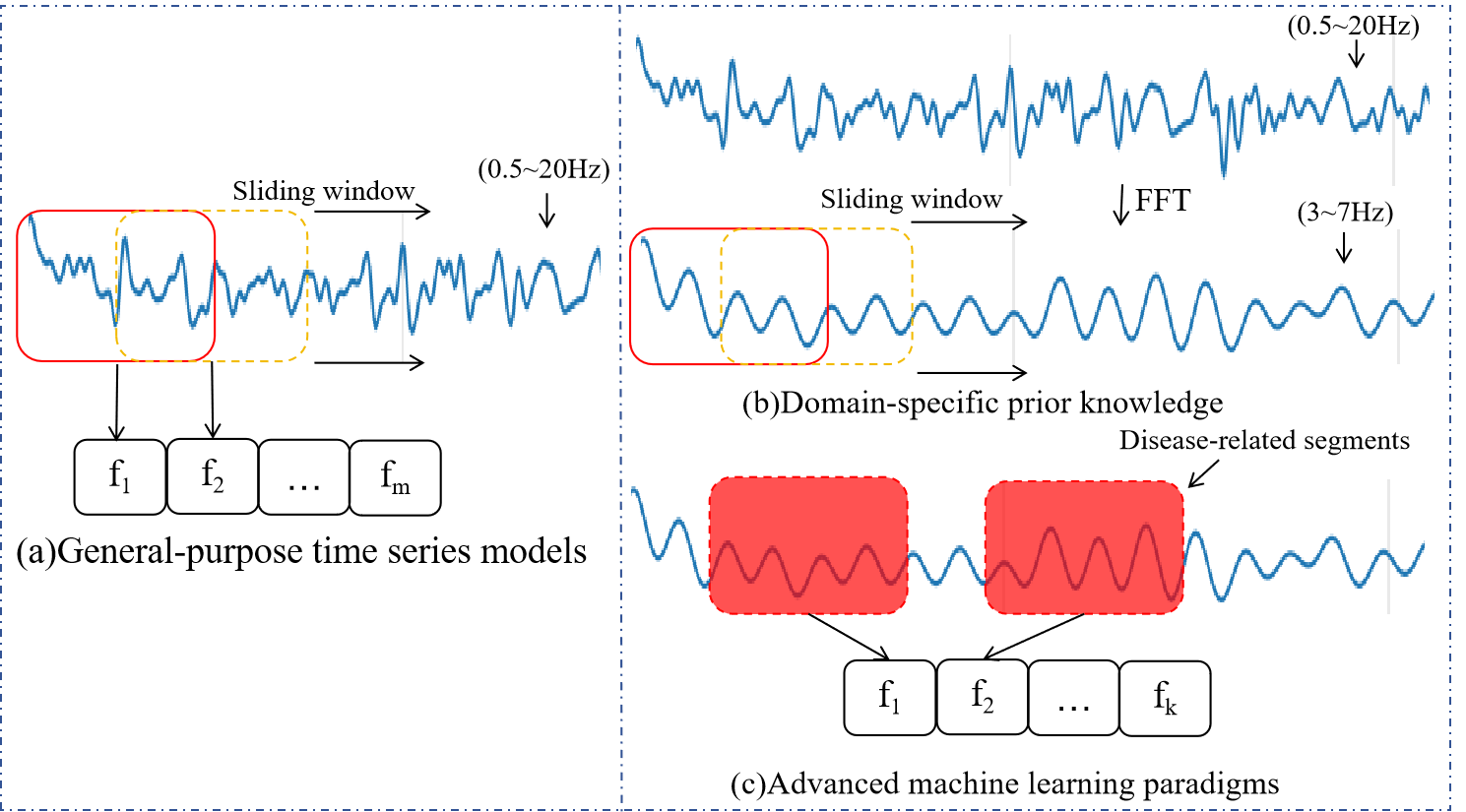}
\caption{Comparison between general-purpose time series models and the proposed MFAM method. (a) General-purpose time series models rely on generic feature extraction, while MFAM integrates (b) domain-specific prior knowledge with (c) advanced machine learning paradigms to address pathological feature specificity and temporal symptom sparsity.}
\label{fig:method_comparison}
\end{figure}
years, with the development of wearable sensor technology and deep learning, automated and objective PD severity assessment using motor signals from daily activities has become a promising frontier in computational health\cite{b3,b4}.

Deep learning-based multivariate time series classification (MTSC) techniques, particularly CNNs, RNNs, and Transformers, have achieved success in general human activity recognition (HAR) tasks\cite{b5}. Deep learning has also shown potential in diagnosing other neurodegenerative diseases\cite{b6}. However, as illustrated in Fig.~\ref{fig:method_comparison}, directly applying these general-purpose time series models to clinical auxiliary diagnosis of specific diseases like PD faces two core challenges. The first challenge is the lack of pathological feature specificity. General-purpose time series models learn universal motion patterns from broad frequency ranges (typically 0.5--20\,Hz), but PD diagnosis highly depends on precise capture of specific pathological signals, such as resting tremor concentrated in narrow frequency bands (typically 3--7\,Hz)\cite{b7}. These subtle pathological signals are often treated as noise by general-purpose time series models, resulting in poor discrimination. The second challenge is the temporal sparsity of symptoms. In long-term monitoring, typical motor symptoms of PD (such as tremor and freezing of gait) are often paroxysmal, appearing only briefly in few segments\cite{b8}. General-purpose time series models often use methods like Global Average Pooling to aggregate features, which severely dilutes key pathological events and reduces model sensitivity.

To tackle these challenges, as shown in Fig.~\ref{fig:method_comparison}, this paper proposes the Multi-scale Frequency-Aware Adversarial Multi-Instance Network (MFAM), which integrates domain-specific prior knowledge with advanced machine learning paradigms. To address insufficient pathological feature specificity, MFAM incorporates a frequency decomposition module that utilizes medical prior knowledge to filter and reconstruct specific frequency bands highly correlated with PD motor symptoms. To address temporal symptom sparsity, MFAM introduces an Attention-based Multi-Instance Learning (Attention-MIL) framework to adaptively focus on the most diagnostically valuable sparse segments. Furthermore, MFAM incorporates a conditional adversarial domain adaptation network to learn robust, individual-agnostic pathological features.

The main contributions of this paper can be summarized as follows:
\begin{itemize}
    \item \textbf{Enhance Specificity:} We introduce a novel frequency decomposition module that integrates medical prior knowledge. Unlike general-purpose time series models that often filter out subtle pathological signals as noise, our module specifically isolates and amplifies the frequency bands highly correlated with PD motor symptoms (e.g., resting tremor), thereby enhancing feature specificity for accurate pathological pattern recognition.
    \item \textbf{Sparse Symptom Detection:} To overcome the challenge of temporal symptom sparsity, where brief pathological events are diluted by general-purpose time series models' aggregation methods, we propose an Attention-based Multi-Instance Learning (Attention-MIL) framework. This framework is specifically designed to identify and adaptively focus on the most diagnostically valuable sparse temporal segments, enabling effective capture of key pathological information.
    \item We validate the effectiveness of MFAM through comprehensive experiments covering a PD versus differential diagnosis task (on the public PADS dataset) and a four-class severity assessment (on a private dataset). The experimental results show that MFAM comprehensively outperforms various state-of-the-art baseline models, demonstrating its superiority and robustness in handling complex clinical application scenarios.
\end{itemize}

\section{Related Work}

\subsection{Deep Learning Models for Time Series Classification}
In recent years, deep learning has become the mainstream paradigm in the MTSC field. Early research primarily revolved around Convolutional Neural Networks (CNNs) and Recurrent Neural Networks (RNNs). CNNs effectively capture local patterns and shape features in time series, such as MCDCNN\cite{b9}. RNNs and their variants, such as Long Short-Term Memory (LSTM) networks, model long-term temporal dependencies through their recurrent structure\cite{b10}.

With the rise of the attention mechanism, Transformer-based models have gained attention for their superior ability to capture long-range dependencies\cite{b11}. Models like PatchTST\cite{b12} segment time series into ``patches'' and use them as input ``tokens'' for the Transformer, achieving state-of-the-art performance in forecasting and classification tasks. Researchers have also proposed hybrid models such as MLSTM-FCN\cite{b10} to combine different model strengths. However, these general-purpose time series models adopt a ``one-size-fits-all'' philosophy, learning universal features from data and often lacking domain-specific optimization. When applied to problems with strong domain-specific priors, such as PD diagnosis, their performance is limited, struggling to distinguish subtle pathological patterns from normal physiological variations.

\subsection{Multi-scale and Frequency Analysis of Time Series}
Human activity signals manifest patterns across multiple temporal scales. To extract multi-scale features, researchers have proposed various strategies. One mainstream approach leverages the Inception network architecture, using multiple convolutional kernels of different sizes to capture features at different receptive fields in parallel\cite{b13,b14,b15}. Another class of methods decomposes signals using signal processing techniques, such as Wavelet Transform that decomposes time series into sub-bands of different frequencies\cite{b16}. Recently, works have utilized the Fourier transform to explore signal periodicity. TimesNet\cite{b17} identifies dominant periods via the Fast Fourier Transform and reshapes the 1D time series into a 2D tensor analyzing intra-period and inter-period variations, while MPTSNet\cite{b18} adopts a similar idea to capture multi-scale periodic patterns. These works inspire using frequency-domain analysis to guide feature learning, but their core objective is discovering all potential periodicities in a data-driven manner, which is effective in general HAR tasks but may introduce pathology-irrelevant noise for PD diagnosis. In contrast, MFAM employs a ``knowledge-driven'' strategy, directly utilizing prior knowledge from neurology to focus on specific frequency bands related to PD motor symptoms, making feature extraction more clinically targeted.

\subsection{Multiple Instance Learning and its Application in the Medical Field}
Multiple Instance Learning (MIL) is a weakly supervised learning paradigm designed for scenarios where labels are imprecise or key information is localized\cite{b19}. In MIL, data is organized into ``bags'' consisting of ``instances.'' Labels are provided only at the bag level, while instance-level labels are unknown. A positive bag is assumed to contain at least one positive instance, whereas a negative bag consists entirely of negative instances.

This paradigm fits medical applications naturally. In digital pathology, a whole-slide image (WSI) is a bag where cell patches are instances, and MIL can locate cancer-containing regions without manual annotation\cite{b20}. In medical time series like ECG or EEG anomaly detection, long monitoring records are bags where only seconds contain key signals, and MIL can automatically locate these sparse but critical abnormal events\cite{b21}. Recent studies have also applied MIL to Parkinson's disease diagnosis, mitigating label noise through weakly supervised learning\cite{b22}.

We adopt this concept to address the temporal sparsity of PD symptoms, treating long-term motion monitoring data as ``bags'' and segmented short-term windows as ``instances.'' Compared to traditional MIL methods, MFAM further introduces an attention mechanism to learn instance weights, enabling the model to automatically focus on the most critical pathological segments and providing interpretability for diagnostic decisions.

\section{Methodology}
\begin{figure*}[!t]
\centering
\includegraphics[width=\textwidth]{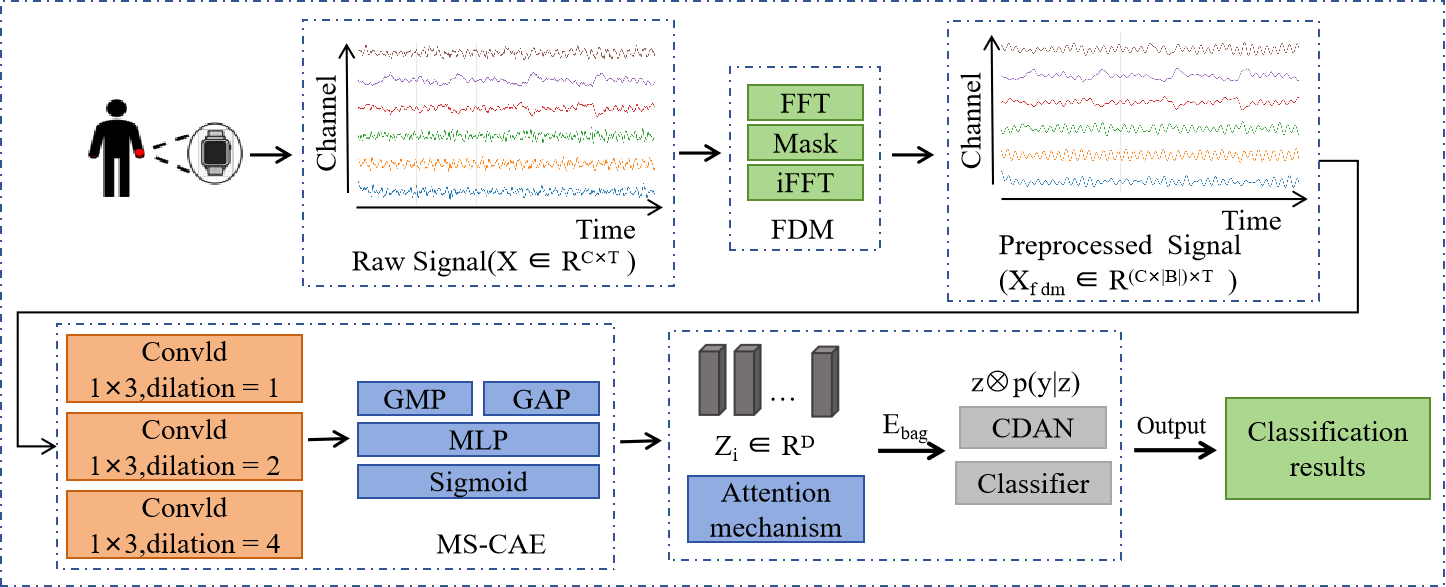}
\caption{Overall architecture of the MFAM model.}
\label{fig:model_architecture}
\end{figure*}

To effectively address the aforementioned challenges, this paper proposes the MFAM model, with an overall architecture depicted in Fig. \ref{fig:model_architecture}. This model constitutes an end-to-end learning framework: an input multivariate time series is first pre-processed by a Frequency Decomposition Module (FDM) in a knowledge-driven manner to isolate specific pathology-related frequency bands; the processed signal is then fed into a Multi-scale Channel Attention Encoder (MS-CAE) for deep feature extraction; subsequently, an attention-based MIL aggregator combines the feature maps of the entire sequence into a key-information-rich bag embedding; finally, this embedding is used by the main classifier for severity level prediction and is simultaneously supervised by a Conditional Adversarial Domain Classifier (CDAN) to learn generalizable representations across individuals. Next, we will first formalize the problem definition and then detail the design of each component.

\subsection{Problem Definition}
This study aims to design a deep learning model for assessing the severity of Parkinson's disease from multivariate time series data collected by wearable sensors. Formally, an input sample is defined as a multivariate time series $\mathbf{X} \in \mathbb{R}^{C \times T}$, where $C$ is the number of sensor channels, and $T$ is the total length of the sequence. The task of the model is to learn a mapping function $f: \mathbb{R}^{C \times T} \rightarrow \{0, 1, \dots, K-1\}$, which can predict the PD severity level label $y$ corresponding to the input sample $\mathbf{X}$, where $K$ is the number of severity levels. The core challenge is that for a PD sample, the signal segments exhibiting pathological features may be sparsely distributed throughout the time series $T$.

\subsection{Frequency Decomposition Module}
The motor symptoms of Parkinson's disease, especially resting tremor, exhibit a distinct concentration of energy in the frequency domain, with energy primarily distributed within specific frequency ranges\cite{b8}. To leverage this domain knowledge, we designed the Frequency Decomposition Module, which aims to precisely isolate the most diagnostically valuable frequency bands.
The specific steps are as follows:
\begin{enumerate}
    \item Fourier Transform: For the input raw signal $\mathbf{X} \in \mathbb{R}^{C \times T}$, we first apply the Fast Fourier Transform (FFT) along the time dimension $T$ to convert it to its frequency domain representation $\hat{\mathbf{X}} \in \mathbb{C}^{C \times F}$, where $F$ is the number of frequency points.
    \item Frequency Band Masking: Based on medical literature, we predefine a set of target frequency bands associated with PD pathology, for example, $B = \{(0.5, 3.0)\,\text{Hz}, (3.0, 7.0)\,\text{Hz}, (7.0, 12.0)\,\text{Hz}\}$. For each frequency band $(f_{low}, f_{high}) \in B$, we create a binary mask $\mathbf{M}_b \in \{0, 1\}^{F}$. This mask has a value of 1 at frequency points within the band's range and 0 elsewhere.
    \item Frequency Domain Filtering: The frequency domain representation $\hat{\mathbf{X}}$ is element-wise multiplied with each band mask $\mathbf{M}_b$ to obtain the frequency domain signal for each band $\hat{\mathbf{X}}_b = \hat{\mathbf{X}} \odot \mathbf{M}_b$.
    \item Inverse Fourier Transform: The Inverse Fast Fourier Transform (IFFT) is applied to each $\hat{\mathbf{X}}_b$ to transform it back into the time domain, yielding a time-domain signal for each frequency band $\mathbf{X}_b \in \mathbb{R}^{C \times T}$.
    \item Channel Concatenation: Finally, we concatenate the time-domain signals $\mathbf{X}_b$ of all frequency bands along the channel dimension $C$ to generate the final output $\mathbf{X}_{fdm} \in \mathbb{R}^{(C \times |B|) \times T}$.
\end{enumerate}
Through this series of operations, the module ``decomposes'' the original signal into multiple parallel feature streams, each carrying specific pathological information, laying a solid foundation for the refined feature extraction of subsequent modules.

\subsection{Multi-scale Channel Attention Encoder (MS-CAE)}
Building on the output $\mathbf{X}_{fdm}$ from the Frequency Decomposition Module, the MS-CAE extracts discriminative temporal representations by integrating multi-scale convolutions with channel attention.

First, the multi-scale convolution branch employs a parallel, Inception-like design, containing three 1D convolution branches with different dilation rates (dilation = 1, 2, 4) and a fixed kernel size of 3, to simultaneously cover short, medium, and long receptive fields. After concatenating the three branches along the channel dimension, they are fused through a 1$\times$1 convolution to obtain an intermediate feature map $\mathbf{X}_{msp} \in \mathbb{R}^{D \times T}$, where $D$ is the hidden dimension.

Subsequently, the channel attention module uses global average pooling (GAP) and global max pooling (GMP) in parallel to aggregate channel-level global information. The results of these two pooling operations are passed through a shared lightweight multi-layer perceptron (MLP, implemented with two 1$\times$1 convolutions), then added together, and passed through a Sigmoid function to generate channel weights $\mathbf{w}_{ch} \in \mathbb{R}^{D \times 1}$. Finally, the channel-weighted output is given by the following equation:
\begin{equation}
\mathbf{X}_{sca} = \mathbf{X}_{msp} \odot \mathbf{w}_{ch}
\end{equation}
where $\odot$ denotes element-wise broadcast multiplication. This integrated encoder not only retains the ability to capture multi-scale context but also emphasizes key feature channels related to diagnosis through an adaptive channel-level mechanism.

\subsection{Attention-based Multi-Instance Learning Aggregator}
This module is a core component of MFAM, designed to tackle the temporal sparsity of PD symptoms. We treat the complete time-series feature map $\mathbf{X}_{sca} \in \mathbb{R}^{D \times T}$ after feature extraction as a ``bag'' and aggregate the final diagnostic evidence from it.
\begin{enumerate}
    \item Instance Construction: We use a sliding window operation to segment the feature map $\mathbf{X}_{sca}$ into $N$ potentially overlapping segments. Each segment, after average pooling, forms an ``instance'' $\mathbf{z}_i \in \mathbb{R}^{D}$, where $i=1, \dots, N$. This results in a bag $\mathcal{Z} = \{\mathbf{z}_1, \dots, \mathbf{z}_N\}$ composed of $N$ instances.
    \item Attention Weight Learning: To evaluate the contribution of each instance to the final diagnosis, we design an attention network. Each instance $\mathbf{z}_i$ is passed through a two-layer fully connected network to compute its attention score $s_i$:
    \begin{equation}
    s_i = \mathbf{w}_2^\top \tanh(\mathbf{W}_1 \mathbf{z}_i + \mathbf{b}_1) + b_2
    \end{equation}
    where $\mathbf{W}_1, \mathbf{b}_1, \mathbf{w}_2, b_2$ are learnable parameters. Then, the Softmax function is applied to the scores of all instances to obtain normalized attention weights $a_i$:
    \begin{equation}
    a_i = \frac{\exp(s_i)}{\sum_{j=1}^{N} \exp(s_j)}
    \end{equation}
    \item Top-K Gating Mechanism: To further enhance the model's focus on key instances, we introduce a Top-K gating mechanism. During training, we only retain the $k$ instances with the highest attention weights and set the weights of the remaining instances to 0, followed by re-normalization of the retained weights. This forces the model to learn a sparser, more discriminative attention distribution.
    \item Weighted Aggregation: Finally, the ``Bag Embedding'' $\mathbf{E}_{bag}$, representing the entire time series, is obtained by a weighted sum of all instances using the attention weights:
    \begin{equation}
    \mathbf{E}_{bag} = \sum_{i=1}^{N} a_i \mathbf{z}_i
    \end{equation}
\end{enumerate}
This bag embedding $\mathbf{E}_{bag} \in \mathbb{R}^{D}$ encapsulates the most relevant information for PD diagnosis from the entire time series and is ultimately fed into a linear classification layer to obtain the final severity prediction.

\subsection{Joint Training and Conditional Adversarial Domain Adaptation}
The bag embedding vector \(\mathbf{E}_{bag}\) generated by the attention-based MIL aggregator is a pivotal component of the model. It is fed into two concurrent branches: a main classifier for predicting PD severity levels, and a conditional adversarial domain classifier (CDAN) for improving generalization.

To enhance the model's generalization ability across different subjects (i.e., domains), we employ conditional adversarial domain classification. Let \(\mathbf{z} \in \mathbb{R}^{D}\) be the bottleneck feature (i.e., \(\mathbf{E}_{bag}\)), and \(\hat{\mathbf{p}}=\mathrm{softmax}(\mathrm{Cls}(\mathbf{z})) \in \mathbb{R}^{K}\) be the class probabilities output by the main classifier. We construct the outer product of the feature and probability
\begin{equation}
\mathbf{h}=\mathrm{vec}(\mathbf{z}\,\hat{\mathbf{p}}^{\top}) \in \mathbb{R}^{D\cdot K}
\end{equation}
as the conditional input to the domain discriminator.

The training objective is to minimize a joint loss function composed of the main classification loss \(\mathcal{L}_{cls}\) and the domain adversarial loss \(\mathcal{L}_{adv}\). Through end-to-end backpropagation, all trainable parameters of the model (mainly in the MS-CAE, MIL aggregator, and CDAN) are optimized simultaneously. In particular, the Gradient Reversal Layer (GRL) in the CDAN ensures the feature extractor learns representations that are both discriminative of disease severity and invariant to individual subjects (domains).

\section{Experiments}

\subsection{Datasets and Evaluation Protocol}
To conduct a comprehensive evaluation of MFAM's performance, we conducted experiments on one private dataset and one public dataset.
\begin{itemize}
    \item PADS Public Dataset\cite{b23}: This is a widely used PD assessment dataset collected via Apple Watch, also containing 6-channel motion signals at a 100\,Hz sampling rate. The dataset includes PD patients and patients with a Differential Diagnosis (DD), the latter referring to patients with Parkinson-like symptoms but who do not have PD. The task on this dataset is a binary classification between PD and DD, which is of great importance for clinical differential diagnosis. In recent years, wearable device-based PD detection and assessment methods have made significant progress\cite{b24,b25,b26}, providing important references for research in this field.
    \item Private Dataset: This dataset was collected by our team and includes a healthy control group (HC) and PD patients at three different severity levels (mild, moderate, severe). Data was collected using a single wrist-worn wearable device, containing 6 channels of tri-axial acceleration and tri-axial angular velocity, with a sampling rate of 100\,Hz. All data were collected while performing a series of standardized motor tasks. The main task on this dataset is a four-class severity assessment.
\end{itemize}
Evaluation Protocol: To rigorously evaluate model generalization, we employed a subject-level (cross-subject) 4-fold cross-validation scheme for both datasets. Specifically, we randomly and evenly divided all participants (not data segments) into 4 groups. In each fold, 3 groups were used as the training set and the remaining 1 group as the test set. This was repeated 4 times to ensure that each subject's data was used for testing exactly once. We report the per-activity Accuracy, Macro-Precision, Macro-Recall, and Macro-F1 score as evaluation metrics.

\subsection{Implementation Settings}
All models were implemented in PyTorch and trained on a single NVIDIA RTX 3090 GPU. We used the Adam optimizer with a learning rate of $5\times10^{-4}$. All models were trained using early stopping based on validation set performance. In the data preprocessing stage, the raw signals were segmented into 1-second windows with a 50\% overlap. For MFAM, the frequency decomposition module selected three frequency bands (0.5--3\,Hz, 3--7\,Hz, 7--12\,Hz) based on medical prior knowledge; the attention-based MIL aggregator used a Top-K gating with a retention rate of 0.3; the adversarial strength of the GRL in the conditional adversarial domain classifier was dynamically scheduled during training.

\begin{table*}[!t]
\caption{PD vs. DD binary classification results on the PADS public dataset}
\label{tab:pads_2cls}
\begin{center}
\resizebox{\textwidth}{!}{%
\begin{tabular}{l|cccc|cccc|cccc|cccc}
\hline
Activity Name & \multicolumn{4}{c|}{LGBM\cite{b16}} & \multicolumn{4}{c|}{LSTMCNN\cite{b28}} & \multicolumn{4}{c|}{ADAMIL\cite{b29}} & \multicolumn{4}{c}{Ours (MFAM)} \\
\cline{2-17}
 & Acc & Pre & Rec & F1 & Acc & Pre & Rec & F1 & Acc & Pre & Rec & F1 & Acc & Pre & Rec & F1 \\
\hline
Relaxed1     & 70.41 & 35.20 & 50.00 & 41.32 & 71.43 & 85.57 & 51.72 & 44.90 & 76.15 & 71.42 & 70.82 & 70.94 & 80.51 & 78.99 & 72.23 & 73.66 \\
Relaxed2     & 72.45 & 73.67 & 54.45 & 50.81 & 77.55 & 73.28 & 70.06 & 71.21 & 77.43 & 73.32 & 70.15 & 71.06 & 77.94 & 74.28 & 69.18 & 70.21 \\
RelaxedTask1 & 76.53 & 78.07 & 62.34 & 63.19 & 76.53 & 81.53 & 61.34 & 61.69 & 77.44 & 73.37 & 68.62 & 69.97 & 78.46 & 75.67 & 68.78 & 70.22 \\
RelaxedTask2 & 76.53 & 87.50 & 60.34 & 60.00 & 72.45 & 73.67 & 54.45 & 50.81 & 76.41 & 71.83 & 67.63 & 68.84 & 78.97 & 78.36 & 67.85 & 69.25 \\
StretchHold  & 71.13 & 61.02 & 52.12 & 47.61 & 74.23 & 86.70 & 55.36 & 52.01 & 77.96 & 73.68 & 71.54 & 72.26 & 78.45 & 74.45 & 69.83 & 70.60 \\
HoldWeight   & 70.41 & 35.20 & 50.00 & 41.32 & 73.47 & 68.42 & 60.17 & 60.61 & 75.64 & 71.18 & 71.64 & 70.97 & 79.74 & 77.84 & 70.95 & 72.23 \\
DrinkGlas    & 70.41 & 35.20 & 50.00 & 41.32 & 77.55 & 75.82 & 66.07 & 67.76 & 79.49 & 76.83 & 70.57 & 72.35 & 79.99 & 76.30 & 72.94 & 74.11 \\
CrossArms    & 71.43 & 64.84 & 54.72 & 52.36 & 74.49 & 74.72 & 58.90 & 58.35 & 79.23 & 76.06 & 71.43 & 72.80 & 81.29 & 79.77 & 73.85 & 75.45 \\
TouchNose    & 73.20 & 68.33 & 56.75 & 55.66 & 77.32 & 73.21 & 67.08 & 68.65 & 79.23 & 76.07 & 70.66 & 72.32 & 79.99 & 78.45 & 71.71 & 73.24 \\
Entrainment1 & 73.20 & 86.32 & 53.57 & 48.74 & 76.29 & 70.97 & 69.54 & 70.15 & 77.96 & 74.62 & 69.22 & 70.62 & 79.49 & 78.19 & 69.24 & 70.76 \\
Entrainment2 & 74.23 & 76.96 & 56.42 & 54.36 & 74.23 & 72.94 & 57.48 & 56.42 & 76.41 & 71.53 & 68.18 & 69.25 & 80.01 & 79.98 & 69.62 & 71.40 \\
\hline
\end{tabular}
}
\end{center}
\end{table*}

\begin{table*}[!t]
\caption{Four-class severity assessment results on the private dataset}
\label{tab:private_4cls}
\begin{center}
\resizebox{\textwidth}{!}{%
\begin{tabular}{l|cccc|cccc|cccc|cccc}
\hline
Activity Name & \multicolumn{4}{c|}{LGBM\cite{b16}} & \multicolumn{4}{c|}{CNNLSTM\cite{b27}} & \multicolumn{4}{c|}{LSTMCNN\cite{b28}} & \multicolumn{4}{c}{Ours (MFAM)} \\
\cline{2-17}
 & Acc & Pre & Rec & F1 & Acc & Pre & Rec & F1 & Acc & Pre & Rec & F1 & Acc & Pre & Rec & F1 \\
\hline
FT    & 54.55 & 54.67 & 39.03 & 39.58 & 48.48 & 36.72 & 27.78 & 20.96 & 48.48 & 28.74 & 33.33 & 28.41 & 63.64 & 51.28 & 58.57 & 55.83 \\
FOC     & 66.67 & 61.01 & 50.00 & 50.98 & 48.48 & 36.72 & 31.25 & 25.96 & 54.55 & 56.71 & 37.92 & 38.21 & 69.70 & 56.31 & 87.05 & 62.78 \\
PSM& 63.64 & 58.33 & 46.25 & 45.60 & 69.70 & 38.53 & 47.22 & 41.96 & 72.73 & 61.04 & 52.22 & 49.17 & 73.33 & 57.66 & 63.12 & 58.33 \\
RHF    & 54.55 & 27.84 & 37.78 & 31.77 & 48.48 & 24.54 & 36.67 & 29.11 & 51.52 & 24.18 & 38.33 & 29.57 & 75.76 & 65.12 & 80.36 & 65.69 \\
LHF    & 66.67 & 64.42 & 46.67 & 46.63 & 63.64 & 52.50 & 46.11 & 45.25 & 66.67 & 86.01 & 51.25 & 55.03 & 69.70 & 48.71 & 59.12 & 51.67 \\
FN-L  & 65.52 & 39.58 & 40.62 & 37.65 & 51.72 & 37.50 & 33.33 & 29.17 & 51.72 & 28.78 & 33.48 & 30.78 & 68.75 & 59.25 & 82.14 & 57.08 \\
FH-R  & 63.64 & 76.50 & 48.47 & 52.50 & 60.61 & 42.45 & 43.33 & 41.87 & 69.70 & 51.67 & 52.78 & 51.32 & 69.70 & 54.47 & 62.50 & 59.44 \\
FRA    & 46.88 & 11.72 & 25.00 & 15.96 & 50.00 & 27.50 & 36.88 & 29.17 & 56.25 & 63.39 & 40.62 & 39.66 & 75.76 & 58.48 & 55.91 & 65.00 \\
WALK     & 45.45 & 11.36 & 25.00 & 15.62 & 54.55 & 62.50 & 41.25 & 40.95 & 54.55 & 62.50 & 41.25 & 40.95 & 66.67 & 60.42 & 84.00 & 60.83 \\
AFC     & 56.67 & 29.00 & 33.33 & 28.66 & 63.33 & 80.61 & 51.88 & 56.13 & 53.33 & 26.04 & 48.21 & 33.77 & 75.86 & 42.81 & 40.22 & 47.22 \\
DRINK   & 61.76 & 46.27 & 49.48 & 47.14 & 73.53 & 66.00 & 53.47 & 53.04 & 76.47 & 66.67 & 59.72 & 60.20 & 81.25 & 73.42 & 74.11 & 76.67 \\
PICK & 57.14 & 43.71 & 45.83 & 41.22 & 60.71 & 29.76 & 35.71 & 31.90 & 64.29 & 50.89 & 49.26 & 47.74 & 71.43 & 51.62 & 62.68 & 50.30 \\
\hline
\end{tabular}
}
\end{center}
\end{table*}

\subsection{Results and Analysis}
We conducted a comprehensive comparison of our proposed MFAM model with a series of representative baseline methods on both the public PADS dataset for the PD versus differential diagnosis (DD) binary classification and the private dataset for four-class severity assessment task. The baselines included a traditional machine learning method LightGBM (LGBM)\cite{b16}, classic hybrid deep learning architectures such as CNNLSTM\cite{b27} and LSTMCNN\cite{b28}, and a highly relevant state-of-the-art model ADAMIL\cite{b29}. 

Tables \ref{tab:pads_2cls} and \ref{tab:private_4cls} present the experimental results of MFAM on the PADS public dataset for PD versus DD binary classification and the private dataset for four-class severity assessment, respectively. The results demonstrate that MFAM achieves the best or near-best performance across almost all activities and evaluation metrics, significantly outperforming general-purpose time series models such as LGBM, CNNLSTM and LSTMCNN, and the domain-relevant state-of-the-art model ADAMIL\cite{b29}. On the PADS dataset, despite the highly similar motor patterns between DD and PD patients, MFAM still demonstrates superior discrimination ability, surpassing all comparative methods in accuracy and F1-score for the vast majority of activities. On the private dataset, MFAM exhibits particularly prominent advantages in activities like ``DrinkWater'' and ``RHandFlip'' that can induce obvious tremors, directly validating the effectiveness of the frequency decomposition module in precisely capturing key features by focusing on pathology-related frequency bands (0.5--3\,Hz, 3--7\,Hz, 7--12\,Hz). For activities with sparse symptom manifestations such as ``WalkBadk'' and ``HandRaise,'' MFAM still performs excellently, benefiting from the attention-based MIL mechanism that can automatically locate and amplify transient pathological segments from long-term signals. This fully proves that the features learned by MFAM are not only sensitive to PD severity but also highly specific, capable of effectively distinguishing the unique pathological patterns of PD from other similar motor disorders.

Synthesizing the results from both experimental sets, we can conclude that, compared to general-purpose time series models relying on handcrafted features or universal patterns, the MFAM framework demonstrates clear performance advantages in both PD severity assessment and differential diagnosis tasks. This advantage stems from MFAM's targeted design—it specifically addresses the two core challenges in PD clinical data, ``insufficient pathological feature specificity'' and ``temporal symptom sparsity,'' by organically combining medical prior knowledge (pathological frequency bands) with advanced machine learning paradigms (attention-based MIL and domain adaptation), enabling it to maintain robust performance even under strict subject-level cross-validation and fully validating the effectiveness and superiority of the proposed method in handling complex clinical time series data.

\subsection{Attention Weight Visualization Analysis}
To intuitively demonstrate how the attention-based MIL module in MFAM effectively addresses the temporal symptom sparsity challenge, we visualized the attention weight distribution of the trained model on real samples. Figure \ref{fig:attention_comparison} presents the attention allocation patterns for two different samples (sample 50 is a PD patient, sample 100 is a DD patient). From the figure, we can clearly observe the following key phenomena:
\begin{figure}[htbp]
\centering
\includegraphics[width=0.48\textwidth]{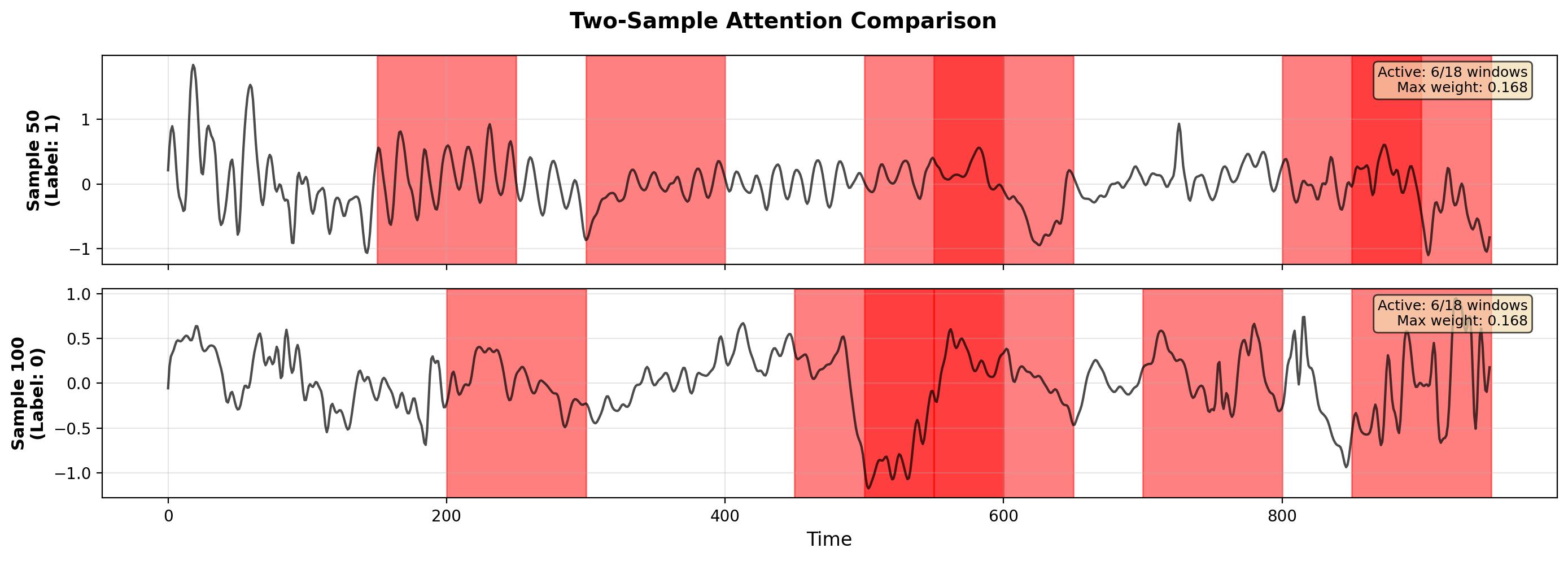}
\caption{Attention weight distribution comparison for two samples. The upper panel shows a PD patient (label 1), and the lower panel shows a DD patient (label 0).}
\label{fig:attention_comparison}
\end{figure}
First, the attention weights exhibit a distinct sparse distribution characteristic. Among the 18 time windows, the model assigns significant weights to only approximately 5--7 windows (red regions), while effectively suppressing the remaining windows (uncolored regions). This sparsity validates the effectiveness of the Top-K gating mechanism, enabling the model to automatically locate and focus on sparse time segments containing critical diagnostic information, thereby avoiding the information dilution problem caused by traditional global average pooling. Second, different sample categories demonstrate differentiated attention patterns. PD patient samples tend to show more concentrated high-weight distributions in specific time windows, which often correspond to the onset periods of typical symptoms such as tremor or bradykinesia. In contrast, DD patient samples exhibit more dispersed attention distributions, reflecting the different characteristics of their motor patterns. This adaptive attention allocation strategy fully demonstrates that the MIL framework can learn temporal feature representations highly correlated with disease categories. Finally, these visualization results provide interpretability support for model decisions. By showing the key time segments that the model ``attends to,'' clinicians can understand the diagnostic basis of the model, thereby enhancing trust in AI-assisted diagnostic systems. This has significant practical implications for applying deep learning models to real clinical scenarios.

\section{Conclusion}
In this paper, we addressed the challenges of using wearable sensors for Parkinson's disease (PD) severity assessment, particularly ``insufficient pathological feature specificity'' and ``temporal symptom sparsity'' that hinder existing time series models. To tackle these challenges, we proposed MFAM, a deep learning framework. The framework's core innovations include: first, a ``knowledge-driven'' frequency decomposition module to precisely extract feature bands highly correlated with PD pathology; second, an attention-based multi-instance learning (MIL) paradigm to effectively capture sparsely distributed key pathological events in long-term monitoring; and finally, a conditional adversarial domain adaptation network to enhance the model's generalization ability across individuals.

We conducted comprehensive validation on a public PD vs. DD binary classification dataset and a private four-class severity level dataset. The results show that MFAM comprehensively surpasses general-purpose time series models in both fine-grained severity assessment and clinical differential diagnosis, fully confirming the effectiveness and superiority of the proposed method in handling complex clinical time series data.

Future work will proceed in three directions: First, exploring extending the current multi-class classification model into a regression model capable of outputting continuous severity scores for more refined disease state quantification. Second, planning to fuse multi-modal data from sensors at different locations\cite{b30} to build a more comprehensive and robust integrated diagnostic system for PD, and exploring domain adaptation techniques to transfer laboratory data to free-living environments. Finally, committed to improving model interpretability by visualizing attention weights to highlight to clinicians the key time segments on which the model bases its diagnostic decisions, thereby enhancing the model's credibility and utility in practical clinical applications.

\vspace{12pt}

\end{document}